\documentclass[final,5p,times,twocolumn]{elsarticle}

\usepackage{booktabs} 
\usepackage{subfig}
\usepackage{amssymb}
\usepackage{lipsum}
\usepackage{natbib}
\newcommand{\etal}{\textit{et al}. }
\usepackage{graphicx}
\usepackage{algorithm}
\usepackage{algpseudocode}%
\usepackage{xcolor}
\usepackage{amsmath}
\usepackage{mathtools}
\usepackage{multirow}
\usepackage{graphics}
\usepackage{enumitem}
\usepackage{CJKutf8}
\usepackage{array}
\usepackage{diagbox}
\usepackage{svg} 
\usepackage{amssymb}
\usepackage{comment}
\usepackage{amsmath}
\usepackage{hyperref}
\usepackage[utf8]{inputenc}
\usepackage{ragged2e}
\usepackage{lettrine}

\makeatother
\biboptions{sort&compress}

\journal{arXiv}

\begin{document}

\begin{frontmatter}



\title{PL-DCP: A Pairwise Learning framework with Domain and Class Prototypes for EEG emotion recognition under unseen target conditions}


\author[1,2]{Guangli Li}
\author[1]{Canbiao Wu}
\author[1]{Zhehao Zhou}
\author[1]{Tuo Sun}
\author[2,3]{Ping Tan}
\author[4,5]{Li Zhang}
\author[4,5,*]{Zhen Liang}

\affiliation[1]{organization={School of Biological Science and Medical Engineering, Hunan University of Technology}, city={Zhuzhou}, country={China}}
\affiliation[2]{organization={Xiangjiang Laboratory}, city={Chandsha}, country={China}}
\affiliation[3]{organization={School of Intelligent Engineering and Intelligent Manufacturing, Hunan University of Technology and Business}, city={Chandsha}, country={China}}
\affiliation[4]{organization={School of Biomedical Engineering, Health Science Center, Shenzhen University}, city={Shenzhen}, country={China}}
\affiliation[5]{organization={Guangdong Provincial Key Laboratory of Biomedical Measurements and Ultrasound Imaging}, city={Shenzhen}, country={China}}
\affiliation[*]{Address correspondence to: janezliang@szu.edu.cn}

\begin{abstract}
    \indent Electroencephalogram (EEG) signals serve as a powerful tool in affective Brain-Computer Interfaces (aBCIs) and play a crucial role in affective computing. In recent years, the introduction of deep learning techniques has significantly advanced the development of aBCIs. However, the current emotion recognition methods based on deep transfer learning face the challenge of the dual dependence of the model on source domain and target domain, As well as being affected by label noise, which seriously affects the performance and generalization ability of the model. To overcome this limitation, we proposes a Pairwise Learning framework with Domain and Category Prototypes for EEG emotion recognition under unseen target conditions (PL-DCP), and integrating concepts of feature disentanglement and prototype inference. Here, the feature disentanglement module extracts and decouples the emotional EEG features to form domain features and class features, and further calculates the dual prototype representation. The Domain-pprototype captures the individual variations across subjects, while the class-prototype captures the cross-individual commonality of emotion categories. In addition, the pairwise learning strategy effectively reduces the noise effect caused by wrong labels. The PL-DCP framework conducts a systematic experimental evaluation on the published datasets SEED, SEED-IV and SEED-V, and the accuracy are 82.88\%, 65.15\% and 61.29\%, respectively. The results show that compared with other State-of-the-Art(SOTA) Methods, the PL-DCP model still achieves slightly better performance than the deep transfer learning method that requires both source and target data, although the target domain  is completely unseen during the training. This work provides an effective and robust potential solution for emotion recognition. The source code is available at \url{https://github.com/WuCB-BCI/PL_DCP}.

\end{abstract}


\end{frontmatter}



\begin{figure*}
\centering
\subfloat{\includegraphics[width=0.9\textwidth]{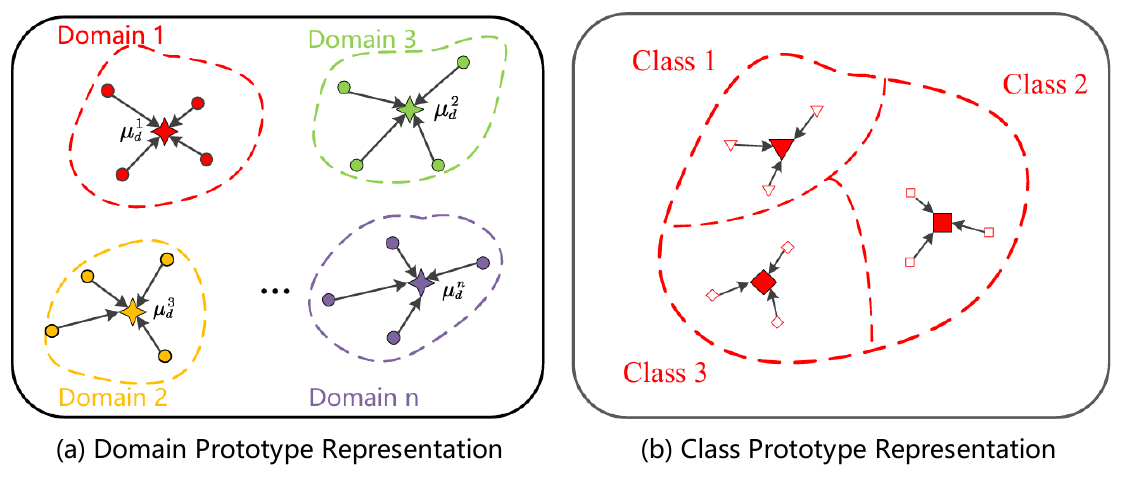}}
\caption{Schematic of the domain prototype and class prototype. (a) Domain Prototype Inference. Colored circles represent domain features, while colored stars represent domain prototypes. Different colors represent different domain prototypes. (b) Class Prototype Inference. Hollow shapes represent class features, and solid shapes represent class prototypes within each domain. Different shapes represent different class prototypes}
\label{fig:get_p}
\end{figure*}

\section{Introduction}
\label{introduction}
    Affective, as a basic psychological state, plays a crucial role in daily life. It not only affects people's feelings, thoughts and behaviors, but also affects people's physical and mental health.\cite{2002Affective} Extensive research has established a strong link between emotional states and mental health disorders.\cite{barrett2001knowing} As a physiological signal, Electroencephalogram (EEG) has the characteristics of high time resolution and objectivity, which provides more direct and objective clues for understanding and evaluating emotional states.\cite{ye2024semi, Ye2023AdaptiveSA} Affective computing is a rapidly developing interdisciplinary research field. Accurately recognizing emotions and providing personalized treatment plans are key challenges in the field of emotion computing, which is crucial for daily life, mental health management, human-computer interaction and so on.\cite{jung2019utilizing}
    \\ \indent In recent years, with the further development of deep learning technology, the field of emotion recognition has also been vigorously developed,\cite{houssein2022human} and has attracted more and more attention from researchers in different fields such as computer science and psychology. It has become a research hotspot in understanding and recognizing human emotions.\cite{hu2020video}\cite{alarcao2017emotions} For example: 
    Feng \etal\cite{Feng_Cheng_Zhao_Deng_Zhang_2022} designed a hybrid model called ST-GCLSTM, which comprises a spatial-graph convolutional network (SGCN) module and an attention-enhanced bi-directional Long Short-Term Memory (LSTM) module, which can be used to extract representative spatial-temporal features from multiple EEG channels.
    Yang \etal\cite{10509712_2024} proposed spectral-spatial attention alignment multi-source domain adaptation ($\mathrm{S^2A^2}$-MSD), which constructs domain attention to represent affective cognition attributes in spatial and spectral domains and utilizes domain consistent loss to align them between domains.
    Zhang \etal\cite{Zhang_Yao_2022} introduced both cascade and parallel convolutional recurrent neural network models for precisely identifying human intended movements and instructions, effectively learning the compositional spatio-temporal representations of raw EEG streams.
    \\ \indent Despite these advances, two key challenges remain that limit the practical application of EEG-based emotion recognition systems. 
    \textbf{(1)} EEG signals are highly personalized. Existing transfer learning models rely heavily on target domain data, which makes the model vulnerable to data preferences, thus reducing its adaptability in the real world environment and increasing the cost of model training. Therefore, we should create emotion recognition models under the condition of unseen target domain to adapt to a large number of individual differences, so as to improve real-world usability. 
    \textbf{(2)} Currently, EEG emotion experiments are basically induced by emotional stimulation material, such as video, pictures, audio, etc. Due to individual physiological factors, subjects may not always respond correctly to emotions. Label noise remains a pressing issue that affects the reliability of the model. It is very important to build an emotion recognition model with the ability to resist label noise to improve the robustness of emotion recognition system. Therefore, addressing these two challenges is crucial to advance the field and achieve more adaptable, accurate and reliable emotion recognition techniques.
    \\ \indent Previous studies have emphasized that EEG signals are highly subject-dependent,\cite{berkhout1968temporal}\cite{paranjape2001electroencephalogram} and that the way individuals perceive and express emotions can vary significantly \cite{gross1997revealing}. These individual differences extend to neural processes involved in emotional regulation, further complicating the task of emotion recognition. \cite{morawetz2024neural} A fundamental assumption in many machine learning and deep learning methods is that training and testing data share the same feature space and follow the same distribution, thus satisfying the independent and identically distributed (IID) condition.\cite{shalev2014understanding}\cite{LeCun_Bengio_Hinton_2015} However, individual variability in EEG signals often disrupts this assumption, leading to substantial performance degradation or even model failure when traditional emotion recognition models are applied to new subjects. This variability presents a significant challenge for the effectiveness and generalizability of existing models, underscoring the need for approaches that can adapt to diverse individual EEG patterns without compromising performance.
    \\ \indent To address the challenges posed by individual variability in EEG signals, a growing number of researchers are employing transfer learning methodologies, which have shown promising results.\cite{zhou2023pr,jayaram2016transfer, zhang2021adaptive,  li2023ms, 9928796,ye2024semi} Transfer learning accommodates variations in domains, tasks and data distributions between training and testing phases. By considering the feature distributions of both the source domain (with labeled data and known distribution) and the target domain (with unlabeled data and unknown distribution), transfer learning approach leverages knowledge from the source domain to improve predictive accuracy in the target domain.\cite{pan2009survey} Given the success of transfer learning in addressing individual differences in EEG signals, most current EEG-based emotion recognition models are developed within a transfer learning framework.\cite{li2018bi, li2019multisource, chen2021ms, gu2022multi} 
    \\ \indent In addition, transfer learning models generally require simultaneous access to both source and target domain data during training, which necessitates retraining the model before it can adapt to new subjects. This requirement significantly increases the practical costs associated with model deployment, particularly when dealing with large datasets or complex models with numerous parameters. Retraining in these cases can be time-consuming and computationally intensive, posing a challenge for efficient deployment and limiting the scalability of these models in real-world, diverse scenarios. Addressing this limitation is crucial for developing more adaptable and cost-effective EEG-based emotion recognition systems. Therefore, addressing the individual differences in EEG signals while avoiding dependence on target domain data is a crucial step for EEG-based emotion recognition models to advance towards practical applications. 
    \\ \indent In order to solve the above problems, We proposes a Pairwise Learning transfer framework based on Domain and Class Prototype (PL-DCP). In this framework, we propose a domain-class dual feature decoupling mechanism for EEG signals, where individual differences are modeled as the superimposed offset of domain features and class features, and fine-grained feature separation is achieved through semantic structure encoding. Domain prototypes (representing the individual differences of subjects, as shown in Fig.\ref{fig:get_p}.(a)) and class prototypes (representing the commonality across subjects' emotional categories, as shown in Fig.\ref{fig:get_p}.(b)) are innovatively constructed, and the sample distribution of the target domain is dynamically calibrated by using prototype similarity matching to alleviate the bottleneck of model generalization caused by individual differences of EEG signals. Alternatively, to reduce the impact of label noise, classification is described as a pairwise learning task that evaluates the relationship between samples and different prototypes. It is worth noting that the target domain data is not needed in the training process, which breaks through the dependence of traditional deep transfer learning on target domain data. Overall, The main contributions are summarized below:
    
    \begin{itemize}
     \item EEG features are represented through a novel dual prototype of domain and class prototypes, enabling individual variability in EEG signals to be conceptualized as feature shifts resulting from interactions between these domains and class prototypes. 
     \item Proposed the concepts of feature disentanglement and prototype reasoning, obtained domain features and class features from EEG signals. And it does not need to touch the target domain data during the training process.
     \item  We rigorously tested model using public databases. Experimental results show that, in the absence of target domain data, PL-DCP still achieves comparable or even better performance than classical deep transfer learning models. In addition, we perform a thorough analysis of our performance on the model and feature visualization to deepen our understanding of the model and results.
    \end{itemize}

\section{Related Work}
\label{sec:related_work}
    \indent To overcome the limitations of the IID assumption, which is challenging to uphold due to significant individual variability in EEG signals, an increasing number of researchers are turning to transfer learning methods. In transfer learning, the labeled training data is source domain, and the unknown test data is target domain. Current transfer learning algorithms for EEG-based emotion recognition can generally be divided into two categories: non-deep transfer learning models and deep transfer learning models.
\subsection{Non-deep transfer learning models}
    \indent To facilitate knowledge transfer between the source and target domains, Pan \etal\cite{TCA2010} proposed the Transfer Component Analysis (TCA) method, which minimizes the maximum mean discrepancy (MMD) to learn transferable components across domains. 
    Fernando \etal\cite{SA2013} introduced the Subspace Alignment (SA) method, which learns a mapping function to align source and target subspaces. The results showed SA could reduce the domain gap and enhance the adaptability of the model. 
    Zheng \etal\cite{zheng2016personalizing} proposed two transfer learning approaches specifically for EEG signals. The first combines TCA with kernel principal component analysis (KPCA) to identify a shared feature space. The second approach, Transductive Parameter Transfer (TPT), constructs multiple classifiers in the source domain and transfers knowledge to the target subject by learning a mapping from distributions to classifier parameters, which are then applied to the target domain. Additionally, 
    Gong \etal\cite{GFK2012} introduced the Geodesic Flow Kernel (GFK), which maps EEG signals to a kernel space that captures domain shifts using geodesic flow. This approach enhances feature alignment by integrating multiple subspaces and identifying domain-invariant directions, thereby supporting more robust cross-domain adaptation.
\subsection{Deep transfer learning models}
    \indent Non-deep transfer models would be limited in complexity and capacity, constraining their ability to fully meet the practical demands of emotion recognition. With advances in deep learning theories and technologies, deep learning-based transfer algorithms have been introduced, offering enhanced model capabilities in terms of performance and generalizability. These algorithms have been widely applied in EEG-based emotion recognition, and most contemporary models now leverage deep transfer learning. For examples, 
    Lin \etal \cite{Yin_Zheng_Hu_Zhang_Cui_2021} proposed a novel emotion recognition method based on a novel deep learning model (ERDL). The model fuses graph convolutional neural network (GCNN) and long-short term memories neural networks (LSTM), to extract graph domain features and temporal features.
    Ye \etal \cite{ye2024semi} introduced a semi-supervised Dual-Stream Self-Attentive Adversarial Graph Contrastive learning framework (DS-AGC) to enhance feature representation in scenarios with limited labeled data. The DS-AGC includes a graph contrastive learning method to extract effective graph-based feature representations from multiple EEG channels. Additionally, it incorporates a self-attentive fusion module for feature fusion, sample selection, and emotion recognition. 
    \\ \indent In recent year, many recent deep transfer learning methods for EEG emotion recognition have been developed based on the Domain-Adversarial Neural Network (DANN\cite{jin2017eeg}) structure, leveraging its capacity for effective domain adaptation. For example, 
    Li \etal \cite{li2018bi} introduced the Bi-Domain Adversarial Neural Network (BiDANN), which accounts for asymmetrical emotional responses in the left and right hemispheres of the brain, leveraging neuroscientific insights to improve emotion recognition performance. 
    Zhang \etal \cite{zhang2019cross} introduced a cross-subject emotion recognition method that utilizes CNNs with DDC. This method constructs an Electrode-Frequency Distribution Map (EFDM) from EEG signals, using a CNN to extract emotion-related features while employing DDC to minimize distribution differences between source and target domains. 
    Gokhale \etal \cite{Gokhale_Anirudh_2022} proposed an adversarial training approach which learns to generate new samples to maximize exposure of the classifier to the attribute-space, which enables deep neural networks to be robust against a wide range of naturally occurring perturbations.
    \\ \indent These deep transfer learning methods has provided valuable insights and strategies for addressing individual differences in EEG signals and achieving significant results in EEG-based emotion recognition. However, a key challenge remains: the dependence of the model on the target domain data during the training process may lead to the model being influenced by the target domain data preference, and increases the practical application cost, which limits the scalability of these models in real-world scenarios.
\subsection{Prototype Learning}
    \indent The core concept of prototype learning is that each class is represented by a prototype (a feature vector that acts as a central, representative feature for that class). Data points belonging to a specific class are clustered around this prototype, enabling classification by evaluating the proximity or similarity of data points to their respective class prototypes. For example, 
    Snell \etal \cite{snell2017prototypica} proposed prototypical networks, which learn a metric space where samples from the same class are clustered around their respective class prototypes. 
    In Pinheiro \etal 's work \cite{pinheiro2018unsupervised}, prototype representations are computed for each class, and target domain images are classified by comparing their feature representations to these prototypes, assigning the label of the most similar prototype. 
    Ji \etal \cite{ji2020few} tackled proposed Semantic-guided Attentive Prototypes Network (SAPNet) framework to address the challenges of extreme imbalance and combinatorial explosion in Human-Object Interaction (HOI) tasks. 
    Liu \etal \cite{Liu2023prototype} developed a refined prototypical contrastive learning network for few-shot learning (RPCL-FSL), which combines contrastive learning with few-shot learning in an end-to-end network for enhanced performance in low-data scenarios. 
    Yang \etal \cite{yang2023two} introduced the Two-Stream Prototypical Learning Network (TSPLN), which simultaneously considers the quality of support images and their relevance to query images, thereby optimizing the learning of class prototypes. 
    These studies show that prototype learning is particularly effective in few-shot learning and unsupervised tasks, significantly improving the robustness of the model and providing a potential approach for the field of affective computing.
    \\ \indent Therefore, in the latest research, Zhou \etal\cite{zhou2023pr} proposed a prototype representation based pairwise learning framework (PRPL), which applies prototype learning to EEG emotion recognition, and interacts sample features with prototype features through bilinear transformation.
    However, PR-PL considers only class prototypes, assuming that source domain data follow a uniform distribution, a limitation that does not align with real-world variability. In addition, PR-PL still relies on source and target data to align sample features during training, which may limit its applicability in real-world scenarios.

\begin{figure*}[ht]
\centering
\subfloat{\includegraphics[width=1\textwidth]{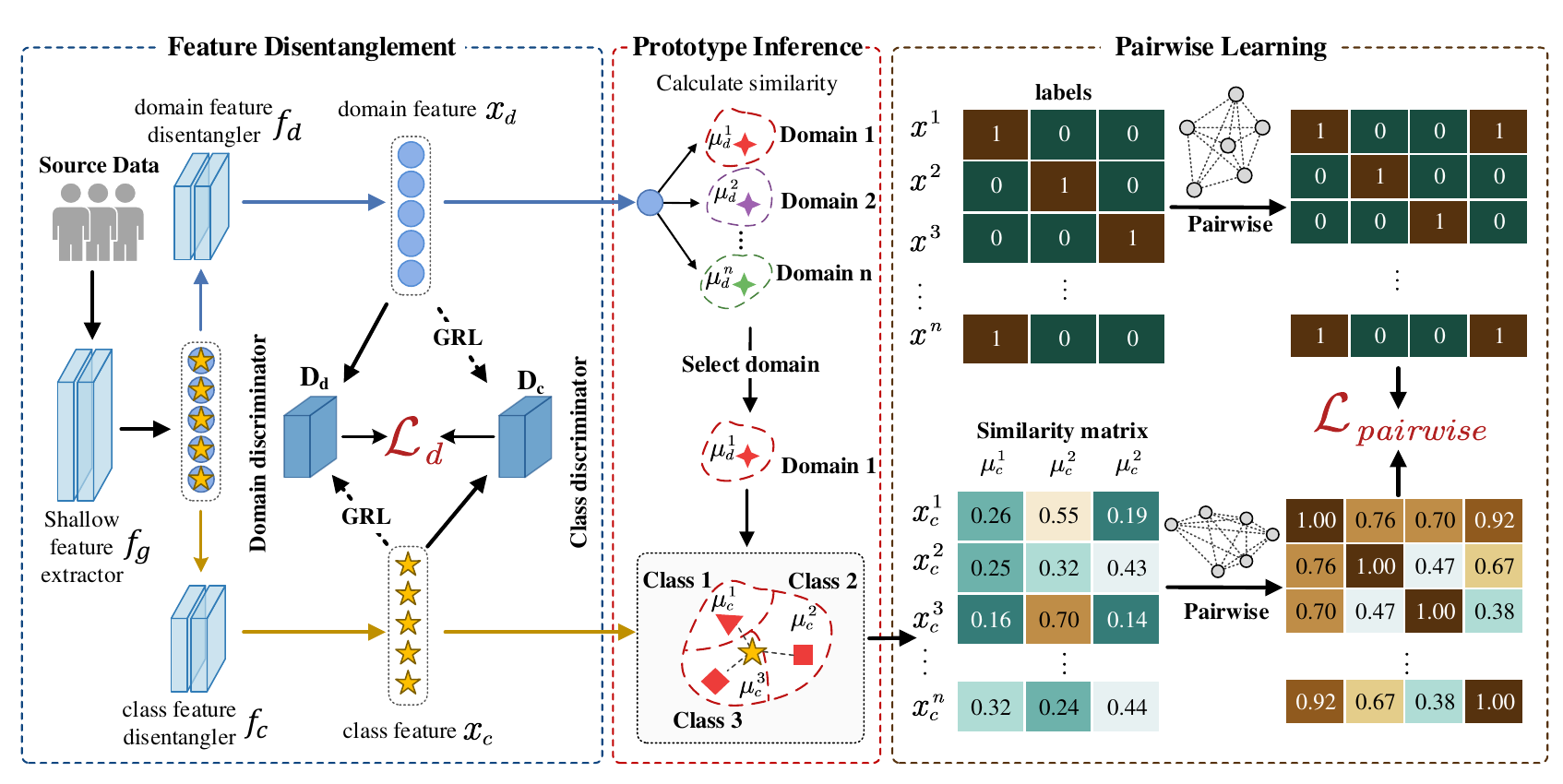}}
\caption{The structural framework of the proposed PL-DCP model. In the feature disentanglement module, we disentangle domain features and class features from shallow EEG features. In the prototype inference module, we obtain the domain and class prototypes for each subject and then assess the similarity between the sample’s domain features and each domain prototype. After selecting the most similar domain, we measure the similarity of the class features of the samples with each class prototype in the domain. In the pairwise learning module, we capture the pairwise relationships between different samples, thereby enhancing the model’s anti-interference to label noise.
}
\label{fig:modelframework}
\end{figure*}

\begin{table}
\centering
\caption{Frequently used notations and descriptions.}
\label{tab:notations}
\begin{tabular}{cc}
\toprule
 Notation                            & Description\\
\midrule
 $\mathbb{S}\backslash \mathbb{T}$   & source\textbackslash{}target domain\\
 $D_d(\cdot) \backslash D_c(\cdot)$  & domain \textbackslash{} discriminator\\
 $f_d(\cdot) \backslash f_c(\cdot)$  & domain\textbackslash{}class feature disentangler\\
 $\mu_d \backslash \mu_c$            &  domain\textbackslash class prototype\\
 $x_d \backslash x_c$                & domain\textbackslash{}class feature\\
 $y_d \backslash y_c $               & domain\textbackslash{}class label\\
 $f(\cdot)$                          & shallow feature extractor\\
 $N_d$                               & the number of subjects in the source domain\\
 $N_c$                               & the number of classes\\
 $S$                                 & bilinear transformation matrix in $h(\cdot)$\\ 
\bottomrule
\end{tabular}
\end{table}
\section{Methodology}
\label{sec:Methodology}
    \indent The source domain is defined as $\mathbb{S}$=\{$S_1$,$S_2$,$S_3$,...,$S_n$$\}^{N_d}_{n=1}$, where $N_d$ denotes the number of subjects in the source domain. For each individual subject in the source domain, we have $S_n$=\{$x^i_n$,$y^i_n$$\}^{N_s}_{i=1}$, where $x^i_n$ denotes the $i$-th sample of $n$-th subject, $y^i_n$ represents the corresponding emotion label, and $N_s$ is the sample size for the $n$-th subject. The target domain is represented as $\mathbb{T}$=\{$x^i_t$,$y^i_t$$\}^{N_t}_{i=1}$, where ${N_t}$ denotes the number of EEG samples in the target domain. For clarity, Tab.\ref{tab:notations} summarizes the commonly used notations.
    \\ \indent Inspired by the work of Peng \etal \cite{peng2019domain} and Cai \etal \cite{cai2019learning}, we hypothesize that EEG features involves two types of deep features: \textit{domain-invariant class features} and \textit{class-invariant domain features}. The domain-invariant class features capture semantic information about the class to which a sample belongs(Such as, negative emotions induce larger LPP amplitudes, and this encoding mechanism is related to individual emotional semantic processing, but does not depend on specific domains.\cite{Schupp_2000}). The class-invariant domain features convey the domain or subject-specific information of the sample(Such as, anatomical differences such as the thickness of the skull and the morphology of the brain regions lead to individual specificity in the spatial distribution of EEG signals. Such characteristics are independent of the emotion class, but affect the baseline of signal amplitude.\cite{Valdes-Hernandez2009}\cite{Haueisen_2002}). Overall, original EEG features can be viewed as an integration of these two types of features.
    \\ \indent The distributional differences in EEG signals across subjects can be attributed to variations in the domain features, causing a shift in the distribution of class features. Since EEG classification adheres to a common standard, the class features from different subjects should ideally occupy the same feature space and follow a consistent distribution. Traditional methods assume that all test data come from a single, unified domain, effectively focusing only on class features while overlooking the presence of domain features. This assumption limits model generalization across subjects. Feature extraction methods such as DANN can be interpreted as an attempt to remove the domain-specific components from sample features, thereby aligning the class features across subjects and improving the generalization performance of the model. 
\subsection{Feature Disentanglement}
    \indent Based on the above theories and assumptions, we consider both domain-specific and class-specific components in EEG feature extraction to improve the robustness and generalization ability across different subjects. To be specific, we start with a shallow feature extractor $f_g$ to obtain shallow features from the EEG samples $x$. Then, we introduce a class feature disentangler $f_c(\cdot)$ and a domain feature disentangler $f_d(\cdot)$ to disentangle the semantic information within these shallow features, resulting in class features $x_{c}$ and domain features $x_{d}$, expressed as:
    \begin{equation}
    \label{Eq:1}
    x_{c} =f_{c} (f_{g}(x))
    \end{equation}
    \begin{equation}
    \label{Eq:2}
    x_{d} =f_{d} (f_{g}(x))
    \end{equation}
    To improve the effectiveness of the disentanglers in separating the two types of features, we introduce a domain discriminator $D_d(\cdot)$ and a class discriminator $D_c(\cdot)$. The domain discriminator is designed to determine the domain of the input features, while the class discriminator ascertains the class of the input features. Our goal is for the domain discriminator to accurately identify the domain of the input when given domain features, while the class discriminator should be unable to identify the class based solely on domain features. This inability to classify based on domain features indicates successful disentanglement, where domain features contain only domain-specific information and no class-related information. Similarly, class features only contain class-related information, without domain-specific information. 
    \\ \indent To achieve this, before the class features are passed into the domain discriminator and the domain features into the class discriminator, they pass through a Gradient Reversal Layer (GRL) to facilitate adversarial training. We use a binary cross-entropy loss function to optimize the discriminators. The output from each discriminator is first passed through a sigmoid layer to obtain probability values, which are then compared to the true labels. This approach converts the multi-class problem into several independent binary classification tasks. The binary cross-entropy loss function is defined as follows:
    \begin{equation}
    \label{Eq:3}
        \mathcal{L} = -\frac{1}{N}\sum_{i=1}^{N}(y_i \cdot \log(z_i) + (1 - y_i) \cdot \log(1 - z_i))
    \end{equation}
    here, $y_i$ represents the true class labels, and $z_i$ denotes the predicted class labels from the discriminator. Specifically, for the class discriminator, the binary cross-entropy loss can be defined as:
    \begin{equation}
    \label{Eq:4}
    \mathcal{L}_{cls}(\theta _{c} )= \mathcal{L}(y_{c}^{i},x_{c}^{i} )+\mathcal{L}(y_{c}^{i},R(x_{d}^{i}))
    \end{equation}
    where $R(\cdot)$ represents the GRL. $y_{c}^{i}$ represents the true class labels of $i$-th sample, $x_{c}^{i}$ represents the class feature of $i$-th sample, and $x_{d}^{i}$ represents the domain feature $i$-th sample. This loss function is iteratively optimized during model training to help the class discriminator accurately distinguish the class labels, ensure that the class features only retain class-related information, and minimize the interference of domain-specific features. Similarly, for the domain discriminator, We define the binary cross-entropy loss function as:
    \begin{equation}
        \label{Eq:5}
        \mathcal{L}_{dom}(\theta _{d} )= \mathcal{L}(y_{d}^{i},x_{d}^{i} )+\mathcal{L}(y_{d}^{i},R(x_{c}^{i}))
    \end{equation}
    here, $y_{d}^{i}$ represents the true domain labels of $i$-th sample. This loss function is optimized to ensure that the domain discriminator accurately identifies the domain-specific information, encouraging the domain features to be disentangled from class-related information. By applying the GRL before the domain features enter the class discriminator, adversarial training is facilitated, further promoting the separation of domain and class features. In the implementation, the shallow feature extractor, domain feature disentangler, and class feature disentangler are all designed as multi-layer perceptrons (MLPs). The disentangled class features $x_c$ and domain features $x_d$ are then utilized in the subsequent prototype inference module. This module learns domain prototypes to represent each domain, and learns class prototypes to capture each class within those domains.
    
\begin{figure}[h]
\centering
\subfloat{\includegraphics[width=0.45\textwidth]{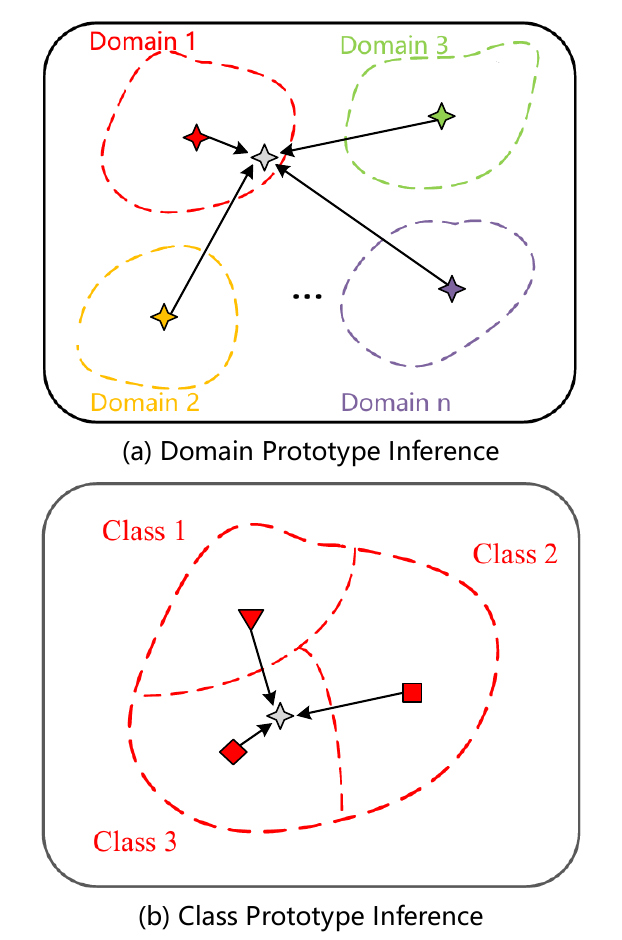}}
\caption{Schematic of domain prototype inference and class prototype inference. Firstly, the prototypes were inferred based on the domain prototypes. . Then, in this domain, the emotion class is inferred based on the class prototype.}
\label{fig:tuili_p}
\end{figure}

\subsection{Prototype Inference}
\label{sec:prototype inference}

    \indent For domain features, we assume that each domain has a prototype representation, which we refer to as the domain prototype. This prototype represents the key characteristics of that domain, with the distribution of domain features centered around it. As shown in Fig.\ref{fig:get_p}(a), For each domain, the domain prototype can be considered the "centroid" of all its domain features. Similarly, for each class within a domain, we derive class prototypes through prototype inference. As shown in Fig.\ref{fig:get_p}(b), the class prototypes capture the essential properties of each class within the domain and serve as the "centroid" of the class features. Both types of prototypes, domain and class, can be computed as the average value of their respective sample features, denoted as $\mu_c$. Specifically, the estimation of domain prototypes for each domain is given by:
    \begin{equation}
    \label{Eq:6}
    \mu _{d} =\frac{1}{\left | X_{n}  \right | }\sum_{x_{d}^{i}\in X_{n} }x_{d}^{i}  
    \end{equation}
    where $X_{n}=\left \{ (x_{d}^{i},y_{d}^{i}  ) \right \}_{i=1}^{|X_{n}|} $ represents the collection of domain features for all samples from the $n$-th subject in the source domain. Here, $|X_{n}|$ denotes the number of samples from this subject, $x_{d}^{i}$ is the the domain feature of the $i$-th sample, and $y_{d}^{i}$ is the corresponding domain label for that sample's domain feature. For data from the same domain, the domain labels $y_{d}^{i}$ are identical. For the class features within a single domain $d^n$, the class prototype is defined as:
    \begin{equation}
    \label{Eq:7}
        \mu _{c}^{d^n,c^*} =\frac{1}{\left | X_{n}^{c^*}  \right | }\sum_{x_{c}^{i}\in X_{n}^{c^*} }x_c^{i}
    \end{equation}
    where $X_{n}^{c^*}=\left \{ (x_{c}^{i},y_{c}^{i}  ) \right \}_{i=1}^{|X_{n}^{c^*}|}$ represents the set of class features for samples that classified as $c^*$ from the $n$-th subject's samples. Here, $|X_{n}^{c^*}|$ denotes the number of samples classified as $c^*$ in the $n$-th subject's data. $x_{c}^{i}$ is the class feature of the $i$-th sample, and $y_{c}^{i}$ is the class label for that sample. In summary, for each sample, we will obtain the corresponding domain prototype $\mu_d$ and class prototypes $[ \mu _{c}^{{d^n},1} ,\mu _{c}^{{d^n},2} ,\cdots ,\mu _{c}^{{d^n},{N_c}} ]$. Here, $N_c$ represents the number of classes. During training, the prototype of each subject is computed based on the features of all its samples. Prototype features are updated iteratively throughout the training process to better capture the feature distribution. Noteworthy, these prototypes are fixed during the testing phase.
    \\ \indent After obtaining the domain prototypes and class prototypes, we proceed with prototype inference. For each sample, as shown in Fig.\ref{fig:tuili_p}.(a), we after feature decoupling and extracting the corresponding domain and class features, we first perform domain prototype inference to identify the most suitable domain. Then,as shown in Fig.\ref{fig:tuili_p}.(b), class prototype inference within the selected domain to determine the class label. Specifically, for the domain feature $x_d^i$, we compare its similarity with each class domain prototype using a bilinear transformation $h(\cdot)$ as:
    \begin{equation}
    \label{Eq:8}
        h(x_d^i,\mu_d)= (x_d^i)^TS\mu_d
    \end{equation}
    where $S \in \mathbb{R}^{d \times d}$ is a trainable, randomly initialized bilinear transformation matrix that is not constrained by positive definiteness or symmetry. The model updates the weights of this bilinear matrix through backpropagation, with the purpose of enhancing the feature representation capability for downstream tasks. 
    We compare the similarity between the sample's domain feature and each domain prototype, defined as:
    \begin{equation}
    \label{Eq:14}
        D_{sim}=softmax([h(x_d^i,\mu_d^1),\dots ,h(x_d^i,\mu_d^{n})])
    \end{equation}
    here, $\mu_d^{n}(n=1,...,N_d)$ represents the domain prototype of the $n$-th subject.The most similar domain $d^*$ is determined based on the $\mu_d^{n}$ corresponding to the maximum value in the vector $D_{sim}$.
    For each training epoch, once the most similar domain $d^*$ for the sample is identified, the class prototypes $\mu_c ^{d^*,k}(k=1,...,N_c)$ for that domain are used to measure the similarity between the sample’s class features and each class prototype. When comparing class features with class prototypes, we use cosine similarity, defined as:
    \begin{equation}
    \label{Eq:9}
        l_i=softmax([d_{cos}(x_c^i,\mu_c^{d^*,1}),\dots ,d_{cos}(x_c^i,\mu_c^{d^*,k})])
    \end{equation}
    where $d_{cos}(\cdot)$ denotes the cosine similarity computation.

\subsection{Pairwise Learning}
\label{sec:Pairwise Learning}
    \indent To solve the label noise problem and enhance the model's resistance to label noise, we employ a pairwise learning strategy to replace pointwise learning. Unlike pointwise learning, pairwise learning takes into account the relationships between pairs of samples, capturing their relative associations through pairwise comparisons. The pairwise loss function used is defined as follows.
    \begin{equation}
    \label{Eq:10}
        \mathcal{L}_{class} =\frac{1}{N_bN_b}\sum_{i,j\in N_b}\mathcal{L}(r_{ij}^s,g(x_c^i,x_c^j;\theta ))
    \end{equation}
    where $\mathcal{L}(\cdot)$ is the binary cross-entropy function, defined in Eq.\ref{Eq:3}. $N_b$ represents the number of samples in a batch. $r_{ij}$ is determined based on the class labels of samples $i$ and sample $j$. For class labels $y_c^i$ and $y_c^j$ of samples $i$ and $j$, if $y_c^i = y_c^j$, then $r_{ij} = 1$; otherwise, $r_{ij} = 0$. The $r_{ij}$ derived from the sample labels enhances the model's stability during the training process as well as its generalization capability. The term $g(x_c^i,x_c^j;\theta)$ represents the similarity measure between the class features of samples $x^i$ and $x^j$, given as:
    \begin{equation}
    \label{eq11}
        g(x_c^i,x_c^j;\theta)=\frac{l_i\cdot l_j}{\left \| l_i \right \|_2 \left \| l_j \right \|_2 } 
    \end{equation}
    Here, $l_i$ and $l_j$ are the feature vectors of the class feature of samples $x^i$ and $x^j$, obtained through prototype inference (Eq.\ref{Eq:9}). The symbol ($\cdot$) represents the dot product operation. The result of $g(x_c^i, x_c^j; \theta)$ falls within the range [0,1], representing the similarity between the two feature vectors $l_i$ and $l_j$. In summary, the objective function for the pairwise learning is defined as follows: 
    \begin{equation}
    \label{eq12}
        \mathcal{L} _{pairwise}(\theta )=\frac{1}{N_bN_b}\sum_{i,j\in N_b}L(r_{ij},  g(x_c^i,x_c^j;\theta ))+\beta \mathcal{R}
    \end{equation}
    Compared to pointwise learning, pairwise learning has a stronger resistance to label noise. Furthermore, a soft regularization term $R$ is introduced to prevent the model from overfitting, with its weight parameter $\beta$ as: 
    \begin{equation}
        \label{eq13}
        \mathcal{R}=||P^{T}P-I {||}_{F} 
    \end{equation}
    where each row of the matrix $P$ represents the domain prototype belonging to a source domain subject, $||\cdot||_F$ denotes the Frobenius norm of the matrix, and $I$ represents the identity matrix.

\section{Experimental Results}
\label{sec:experiment}

\subsection{Dataset and Data Preprocessing}
    \indent We validate the proposed PL-DCP using the widely recognized public databases SEED \cite{7104132}, SEED-IV \cite{zheng2018emotionmeter} and SEED-V\cite{Li_2021_eeg}. The SEED dataset includes 15 subjects, each participating in three experimental sessions conducted on different dates, with each session containing 15 trials. During these sessions, video clips were shown to evoke emotional responses (negative, neutral, and positive) while EEG signals were simultaneously recorded. For the SEED-IV dataset, it includes 15 subjects, each participating in three sessions held on different dates, with each session consisting of 24 trials. In this dataset, video clips were used to induce emotions of happiness, sadness, calmness, and fear in the subjects. In the SEED-V database, a total of 16 subjects were participated, with each completing three sessions on different dates. Each session comprised 15 trials and include five emotions: happiness, neutral, sadness, disgust and fear.
    \\ \indent The acquired EEG signals undergo preprocessing as follows. First, the EEG signals are downsampled to a 200 Hz sampling rate, and noise is manually removed, such as electromyography (EMG) and electrooculography (EOG). The denoised data is then filtered using a bandpass filter with a range of 0.3 Hz to 50 Hz. For each experiment, the signals are segmented using a 1-second window, and differential entropy (DE) features, representing the logarithmic energy spectrum of specific frequency bands, are extracted based on five frequency bands: Delta (1-3 Hz), Theta (4-7 Hz), Alpha (8-12Hz), Beta (14-30Hz), and Gamma (31-50Hz), resulting in 310 features for each EEG segment (5 frequency bands × 62 channels). Finally, a Linear Dynamic System (LDS) is applied to smooth all obtained features, leveraging the temporal dependency of emotional changes to filter out EEG components unrelated to emotions and those contaminated by noise.\cite{5627125} The EEG preprocessing procedure adheres to the same standards as previous studies to enable fair comparisons with models presented in previous literature. All models are executed under the following configuration: NVIDIA GeForce RTX 3090, CUDA=11.6, PyTorch =1.12.1.

\subsection{Experiment Protocols}
    \indent To thoroughly evaluate the model's performance and enable a comprehensive comparison with existing methods, we adopt two different cross-validation protocols. \textbf{(1) Cross-Subject Single-Session Leave-One-Subject-Out Cross-Validation.} This is the most widely used validation method in EEG-based emotion recognition tasks. In this approach, data from a single session of one subject in the dataset is designated as the target, while data from single sessions of the remaining subjects serve as the source. To ensure consistency with other studies, we use only the first session for the cross-subject single-session cross-validation. \textbf{(2) Cross-Subject Cross-Session Leave-One-Subject-Out Cross-Validation.} To more closely simulate practical application scenarios, we also assess the model's performance for unknown subjects and unknown sessions. Similar to the previous method, all session data from one subject in the dataset is assigned as the target domain, while data from all sessions of the remaining subjects serve as the source domain.
    
\begin{table}[t]
\begin{center}
\caption{Cross-subject single-session leave-one-subject-out cross-validation results on SEED dataset, expressed as (Mean-Accuracy$\% \pm$Standard-Deviation\%). Here, '*' indicates the results are obtained by our own implementation.}
\label{tab:seedCompare}
\scalebox{0.9}{
\color{black}
\begin{tabular}{lc|lc}
\toprule
Methods                     & $P_{acc}(\%)$ & 
Methods                     & $P_{acc}(\%)$ \\
\midrule
\multicolumn{4}{c}{\textit{Traditional machine learning methods}} \\ 
\midrule
KNN*\cite{KNN1982}          & 55.26±12.43 &
KPCA*\cite{KPCA1999}        & 48.07±09.97\\
SVM*\cite{2017baseline}     & 70.62±09.02 & 
SA*\cite{SA2013}            & 59.73±05.40 \\
TCA*\cite{TCA2010}          & 58.12±09.52 & 
CORAL*\cite{CORAL2016}      & 71.48±11.57 \\
GFK*\cite{GFK2012}          & 56.71±12.29 & 
RF*\cite{breiman2001random} & 62.78±06.60 \\
\midrule
\multicolumn{4}{c}{\textit{Deep learning methods}}\\ 
\midrule
DAN*\cite{He2018DAN}        & 82.54±09.25  & 
DANN*\cite{LiJDA2020}       & 81.57±07.21  \\
DCORAL*\cite{Dcoral2016}    & 82.90±06.97  & 
DDC*\cite{DDC2014}          & 75.42±10.15  \\
DGCNN\cite{song2018eeg}     & 79.95±09.02  & 
MMD \cite{2013Equivalence}  & 80.88/10.10  \\
BiDANN\cite{li2018bi}       & 83.28±09.60  & 
R2G-STNN\cite{li2019regional}  & 84.16±07.10 \\
EFDMs\cite{zhang2019cross}  & 78.40±06.76 & 
MS-MDA*\cite{chen2021ms}    & 77.65±11.32 \\
\midrule
\multicolumn{3}{l}{\textbf{PL-DCP}} & \textbf{82.88±05.23}\\
\bottomrule
\end{tabular}
}
\end{center}
\end{table}

\begin{table}[t]
\begin{center}
\caption{Cross-subject single-session leave-one-subject-out cross-validation results on SEED-IV dataset, expressed as (Mean-Accuracy$\% \pm$Standard-Deviation\%). Here, '*' indicates the results are obtained by our own implementation.}
\label{tab:seedivCompare}
\scalebox{0.9}{
\color{black}
\begin{tabular}{lc|lc}
\toprule
Methods                     & $P_{acc}(\%)$ & 
Methods                     & $P_{acc}(\%)$ \\
\midrule
\multicolumn{4}{c}{\textit{Traditional machine learning methods}} \\ 
\midrule
KNN*\cite{KNN1982}          & 41.77±09.53 & 
KPCA*\cite{KPCA1999}        & 29.25±09.73 \\
SVM*\cite{2017baseline}     & 50.50±12.03 & 
SA*\cite{SA2013}            & 34.74±05.29 \\
TCA*\cite{TCA2010}          & 44.11±10.76 & 
CORAL*\cite{CORAL2016}      & 48.14±10.38 \\
GFK*\cite{GFK2012}          & 43.10±09.77 & 
RF*\cite{breiman2001random} & 52.67±13.85 \\
\midrule
\multicolumn{4}{c}{\textit{Deep learning methods}}\\ 
\midrule
DAN*\cite{He2018DAN}        & 59.27±14.45  & 
DANN*\cite{LiJDA2020}       & 57.16±12.61  \\
DCORAL*\cite{Dcoral2016} & 56.05±15.60  & 
DDC*\cite{DDC2014}          & 58.02±15.14  \\
MS-MDA*\cite{chen2021ms}    & 57.36±11.76  & 
MMD\cite{2013Equivalence}   & 59.34±05.48  \\
\midrule
\multicolumn{3}{l}{\textbf{PL-DCP}} & \textbf{65.15±10.34}\\
\bottomrule
\end{tabular}
}
\end{center}
\end{table}

\begin{table}[t]
\begin{center}
\caption{Cross-subject single-session leave-one-subject-out cross-validation results on SEED-V dataset, expressed as (Mean-Accuracy$\% \pm$Standard-Deviation\%). Here, '*' indicates the results are obtained by our own implementation.}
\label{tab:seedvCompare}
\scalebox{0.90}{
\color{black}
\begin{tabular}{lc|lc}
\toprule
Methods                     & $P_{acc}(\%)$ & 
Methods                     & $P_{acc}(\%)$ \\
\midrule
\multicolumn{4}{c}{\textit{Traditional machine learning methods}} \\ 
\midrule
KNN*\cite{KNN1982}          & 35.73±07.98 & 
KPCA*\cite{KPCA1999}        & 35.47±09.39 \\
SVM*\cite{2017baseline}     & 53.14±10.10 & 
SA*\cite{SA2013}            & 36.06±11.55 \\
TCA*\cite{TCA2010}          & 37.57±13.47 & 
CORAL*\cite{CORAL2016}      & 55.18±07.42 \\
GFK*\cite{GFK2012}          & 38.32±10.11 & 
RF*\cite{breiman2001random} & 42.29±16.02 \\
\midrule
\multicolumn{4}{c}{\textit{Deep learning methods}}\\ 
\midrule
DAN*\cite{He2018DAN}        & 59.36±16.83  & 
DANN*\cite{LiJDA2020}       & 56.28±16.25  \\
DCORAL*\cite{Dcoral2016}    & 56.26±14.56  & 
DDC*\cite{DDC2014}          & 56.54±18.35  \\
\midrule
\multicolumn{3}{l}{\textbf{PL-DCP}} & \textbf{61.29±09.61}\\
\bottomrule
\end{tabular}
}
\end{center}
\end{table}

\begin{table}[t]
\begin{center}
\caption{Cross-subject cross-session leave-one-subject-out cross-validation results on SEED dataset, expressed as (Mean-Accuracy$\% \pm$Standard-Deviation\%). Here, '*' indicates the results are obtained by our own implementation.}
\label{tab:seedfullycross}
\scalebox{0.9}{
\color{black}
\begin{tabular}{lc|lc}
\toprule
Methods                     & $P_{acc}(\%)$ & 
Methods                     & $P_{acc}(\%)$ \\
\midrule
\multicolumn{4}{c}{\textit{Traditional machine learning methods}} \\ 
\midrule
KNN*\cite{KNN1982}          & 60.18±08.10 & 
KPCA*\cite{KPCA1999}        & 72.56±06.41 \\
SVM*\cite{2017baseline}     & 68.01±07.88 & 
SA*\cite{SA2013}            & 57.47±10.01 \\
TCA*\cite{TCA2010}          & 63.63±06.40 & 
CORAL*\cite{CORAL2016}      & 55.18±7.42 \\
GFK*\cite{GFK2012}          & 60.75±08.32 & 
RF*\cite{breiman2001random} & 72.78±06.60 \\
\midrule
\multicolumn{4}{c}{\textit{Deep learning methods}}\\ 
\midrule
DAN*\cite{He2018DAN}        & 78.12±05.47 & 
DANN*\cite{LiJDA2020}       & 78.42±07.57 \\
DCORAL*\cite{Dcoral2016}    & 77.36±06.27 &
DDC*\cite{DDC2014}          & 73.22±05.48 \\
\midrule
\multicolumn{3}{l}{\textbf{PL-DCP}} & \textbf{79.34±06.34}\\
\bottomrule
\end{tabular}
}
\end{center}
\end{table}

\begin{table}[t]
\begin{center}
\caption{Cross-subject cross-session leave-one-subject-out cross-validation results on SEED-IV dataset, expressed as (Mean-Accuracy$\% \pm$Standard-Deviation\%). Here, '*' indicates the results are obtained by our own implementation.}
\label{tab:seedivfullycross}
\scalebox{0.9}{
\color{black}
\begin{tabular}{lc|lc}
\toprule
Methods                     & $P_{acc}(\%)$ & 
Methods                     & $P_{acc}(\%)$ \\
\midrule
\multicolumn{4}{c}{\textit{Traditional machine learning methods}} \\ 
\midrule
KNN*\cite{KNN1982}          & 40.06±04.98 & 
KPCA*\cite{KPCA1999}        & 47.79±07.85 \\
SVM*\cite{2017baseline}     & 48.36±07.51 & 
SA*\cite{SA2013}            & 40.34±05.85 \\
TCA*\cite{TCA2010}          & 43.01±07.13 & 
CORAL*\cite{CORAL2016}      & 50.01±07.93 \\
GFK*\cite{GFK2012}          & 43.48±06.27 & 
RF*\cite{breiman2001random} & 48.16±09.43 \\
\midrule
\multicolumn{4}{c}{\textit{Deep learning methods}}\\ 
\midrule
DAN*\cite{He2018DAN}        & 60.95±09.34 & 
DANN*\cite{LiJDA2020}       & 61.44±11.66 \\
DCORAL*\cite{Dcoral2016}    & 59.96±09.03 & 
DDC*\cite{DDC2014}          & 54.76±09.02 \\
\midrule
\multicolumn{3}{l}{\textbf{PL-DCP}} & \textbf{63.16±09.03}\\
\bottomrule
\end{tabular}
}
\end{center}
\end{table}

\begin{table}[t]
\begin{center}
\caption{Cross-subject cross-session leave-one-subject-out cross-validation results on SEED-V dataset, expressed as (Mean-Accuracy$\% \pm$Standard-Deviation\%). Here, '*' indicates the results are obtained by our own implementation.}
\label{tab:seedvfullycross}
\scalebox{0.9}{
\color{black}
\begin{tabular}{lc|lc}
\toprule
Methods                     & $P_{acc}(\%)$ & 
Methods                     & $P_{acc}(\%)$ \\
\midrule
\multicolumn{4}{c}{\textit{Traditional machine learning methods}} \\ 
\midrule
KNN*\cite{KNN1982}          & 35.28±07.57 & 
KPCA*\cite{KPCA1999}        & 39.68±11.28 \\
SVM*\cite{2017baseline}     & 41.20±10.76 & 
SA*\cite{SA2013}            & 31.87±09.87 \\
TCA*\cite{TCA2010}          & 37.68±08.40 & 
CORAL*\cite{CORAL2016}      & 54.68±07.44 \\
GFK*\cite{GFK2012}          & 37.89±09.84 & 
RF*\cite{breiman2001random} & 43.63±11.38 \\
\midrule
\multicolumn{4}{c}{\textit{Deep learning methods}}\\ 
\midrule
DAN*\cite{He2018DAN}        & 54.27±10.42 & 
DANN*\cite{LiJDA2020}       & 52.83±13.90 \\
DCORAL*\cite{Dcoral2016}    & 52.23±12.76 & 
DDC*\cite{DDC2014}          & 43.89±11.84 \\
\midrule
\multicolumn{3}{l}{\textbf{PL-DCP}} & \textbf{57.53±09.72}\\
\bottomrule
\end{tabular}
}
\end{center}
\end{table}


\subsection{Cross-Subject Single-Session Leave-One-Subject-Out Cross-Validation.}
    \indent The experimental results of PL-DCP on the SEED database are shown in Tab.\ref{tab:seedCompare}, our proposed PL-DCP model demonstrates a significant performance advantage over both traditional machine learning and non-deep transfer learning methods. Compared to the best-performing non-deep transfer learning method (CORAL), our model shows an improvement of 11.40\% (CORAL: 71.48\%; PL-DCP: 82.88\%). Notably, while other deep transfer learning methods incorporate target domain data during training, our model trains without using any target domain data. Despite this, our method achieves results comparable to, and frequently surpassing, other deep transfer learning approaches that use target domain data, with a 7.46\% improvement over DDC and a 5.23\% improvement over MS-MDA. The results on the SEED-IV dataset are shown in Tab.\ref{tab:seedivCompare}, PL-DCP also achieves superior results without target domain data in training. Compared to the highest-performing traditional machine learning method (RF: 52.67\%), and deep learning methods that utilized target domain data during training (MMD: 59.34\%), PL-DCP achieves an accuracy of 65.15\% ± 10.34\%, outperforming all other methods in both the traditional and deep learning methods. The results on the SEED-V database are shown in Tab.\ref{tab:seedvCompare}. The results of PL-DCP are better than those of traditional machine learning methods. The PL-DCP model achieves an accuracy of 61.29\%, which is 1.93\% higher than the suboptimal method (DAN: 59.36\%). 
    \\ \indent All these results prove that PL-DCP achieves superior performance even without relying on target domain data, and has strong robustness and generalization ability.


\begin{figure*}[t]
\centering
\includegraphics[width=1\textwidth]{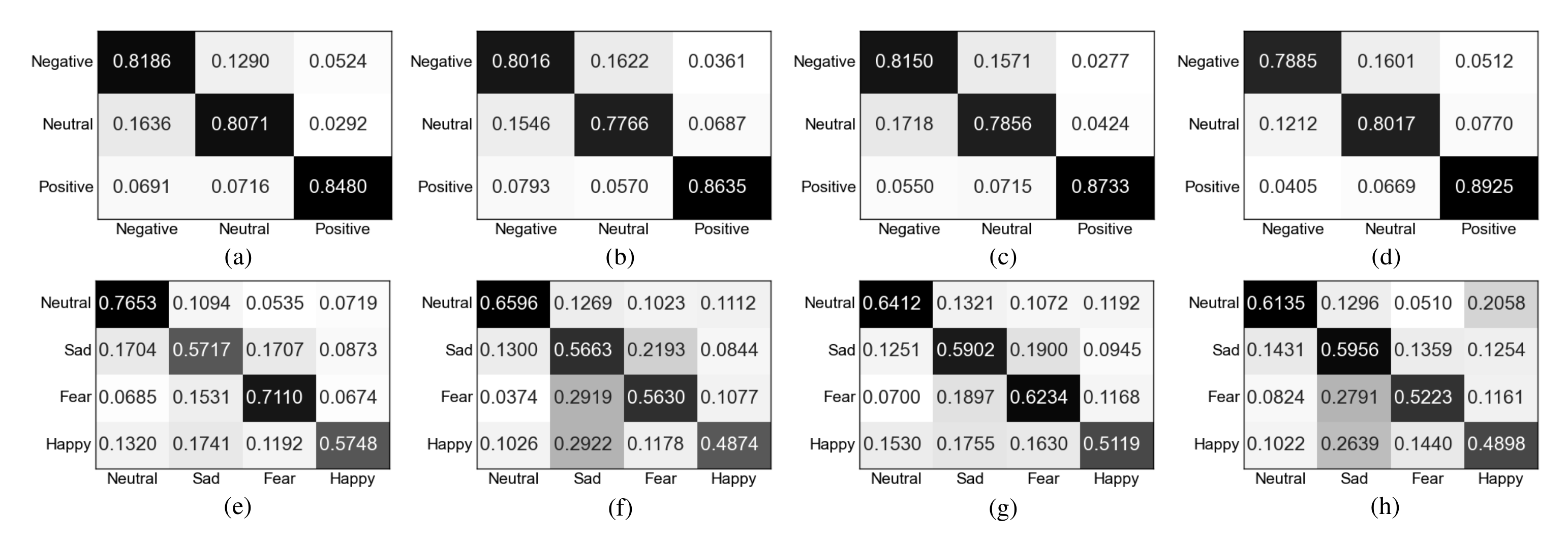}
\caption{Confusion matrices of different model settings under cross-subject single-session leave-one-subject-out cross-validation. The Seed database contains three emotion categories: negative, neutral and positive. Among them, (a) PL-DCP; (b) DANN; (c) DAN; (d) DCORAL. the Seed-IV database contains four emotion categories: happy, sad, calm and fear. Among them, (e) PL-DCP; (f) DANN; (g) DAN; (h) DCORAL.}
\label{fig:confusion_matrix_III_IV}
\end{figure*}

\subsection{Cross-Subject Cross-Session Leave-One-Subject-Out Cross-Validation.}
    \indent Compared to cross-subject single-session, cross-subject cross-session not only accounts for variability among subjects but also incorporates differences across sessions. In EEG-based emotion recognition tasks, this evaluation scheme presents the greatest challenge to the model's effectiveness. The test results of the PL-DCP model in the SEED database are shown in Tab.\ref{tab:seedfullycross}. The PL-DCP model achieves an accuracy of 79.34\%, which is significantly higher than that of traditional machine learning methods and deep transfer learning methods. Compared with the suboptimal model (DANN: 78.24\%), the performance of the PL-DCP model is improved by 0.92\%. As shown in Tab.\ref{tab:seedivfullycross}, the PL-DCP model achieves the best emotion recognition performance in the SEED-IV database, with an accuracy of 63.16\%, which is significantly improved compared with traditional machine learning methods and deep learning methods. Compared with the suboptimal model (DANN: 61.44\%), the performance of PL-DCP is improved by 1.72\%. The test results of the PL-DCP model on the SEED-V database are shown in Tab.\ref{tab:seedvfullycross}, the classification performance of 57.53\% ± 09.72\% was achieved in the five EEG emotion categories of the SEED-V database, compared with the suboptimal performance (CORAL: 54.68\%), the accuracy is improved by 2.85\%. It is worth noting that although the target domain data is completely unknown during training, the PL-DCP model still achieves slightly better performance than the deep transfer learning method that requires both source and target data.
    \\ \indent All these results suggest that the proposed PL-DCP can maintain robust performance independently of target domain data, effectively handling the challenges posed by inter-subject and inter-session variability in EEG-based emotion recognition tasks, which demonstrates the model’s strong validity and generalization capabilities.

\begin{figure*}[t]
\centering
\includegraphics[width=1\textwidth]{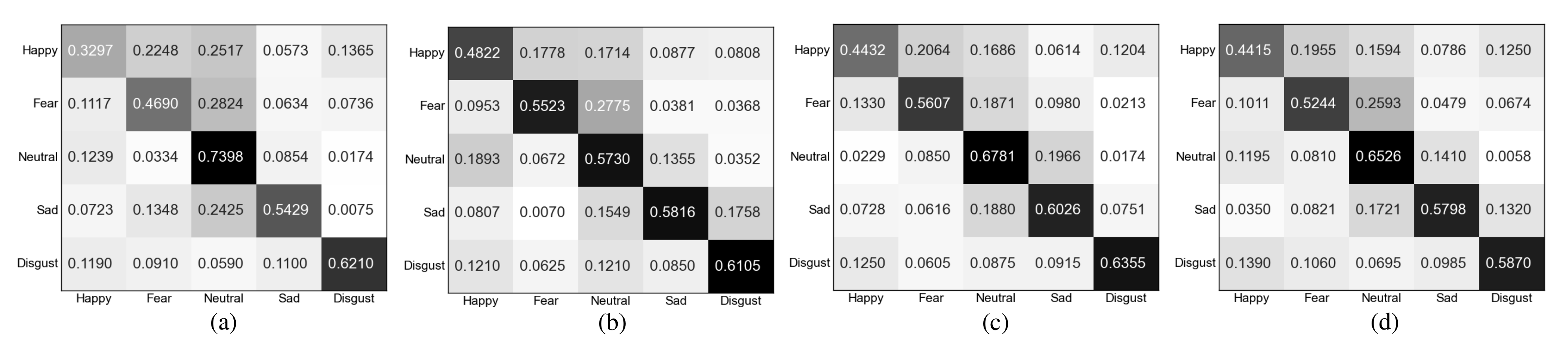}
\caption{Confusion matrices of different model settings under cross-subject single-session leave-one-subject-out cross-validation. The Seed-V database contains five emotion categories: happiness, neutral, sadness, disgust and fear. Among them, (a) PL-DCP; (b) DANN; (c) DAN; (d) DCORAL.}
\label{fig:confusion_matrix_V}
\end{figure*}
\begin{figure*}[ht]
\begin{center}
\includegraphics[width=1\textwidth]{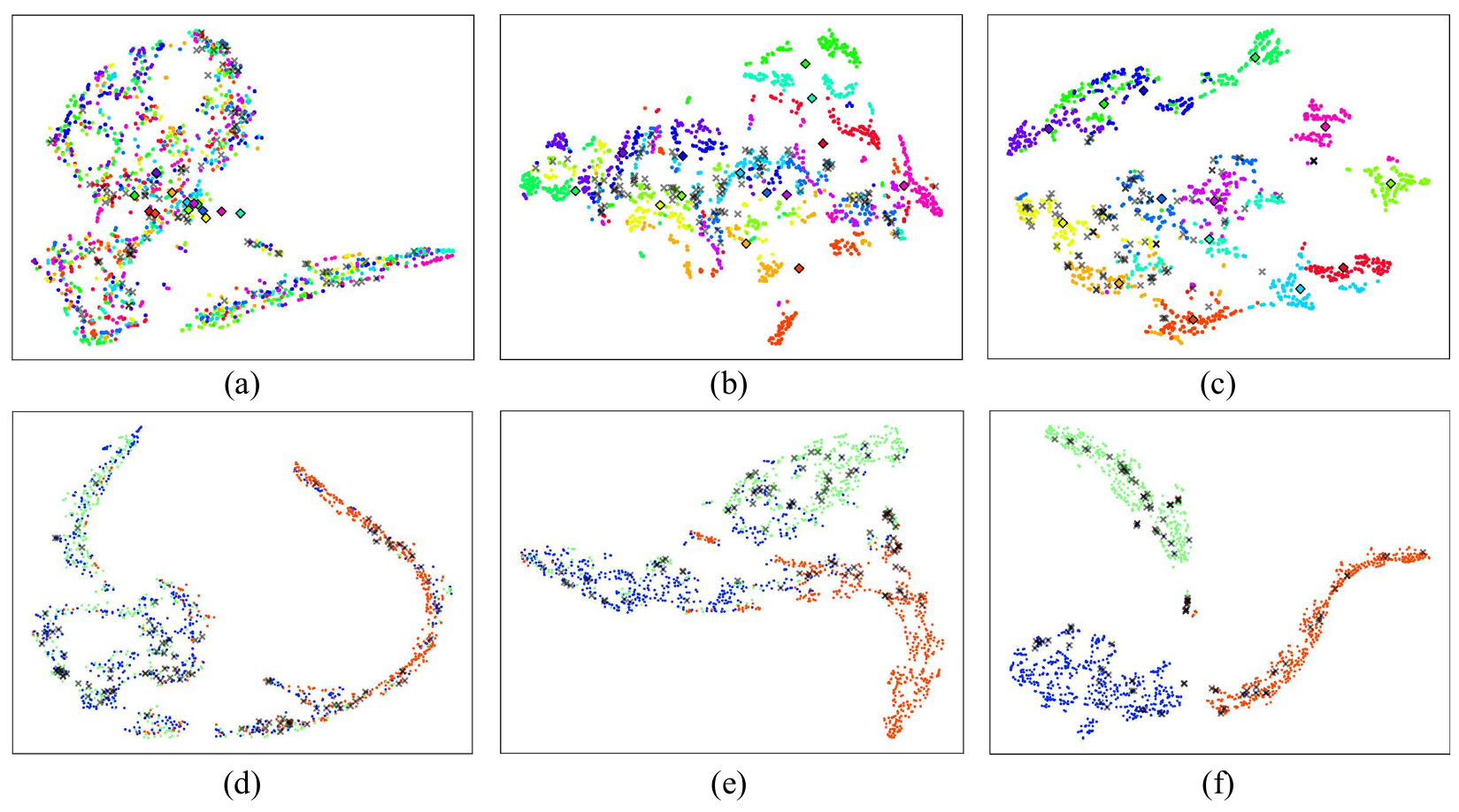}
\end{center}
\caption{Visualization of domain and class features at different training stages. (a)$\sim$(c) show the domain features, where the diamond shape ($\diamond$) represents the domain prototype of each domain, different colors represent the domain features from different domains, the same color represents the domain features from the same domain, and the target domain features are represented as translucent black crosses (×). (d)$\sim$(f) show the class features, where different colors represent different class features. Here, the first column represents the initial stage of training, the second column represents the results after 50 epochs of training, the third column represents the results end of training.}
\label{fig:fea_tsne}
\end{figure*}

    
\subsection{Confusion Matrix}
    \indent The overall accuracy reflects the global performance of the model, but in the practical application of emotional brain-computer interface, the misjudgment of a specific emotion class may cause completely different interaction consequences. In order to qualitatively evaluate the performance of the proposed model on different sentiment categories, we visualize the results of leave-one-out cross-validation on the SEED, SEED-IV and SEED-V datasets using confusion matrices, and compare the results with deep transfer learning methods. To measure the recognition performance of the model for different emotions.
    \\ \indent On the SEED dataset, as shown in Fig.\ref{fig:confusion_matrix_III_IV}.(a)$\sim$(d), all models have the best performance in recognizing positive emotion, which are 81.86\%, 80.16\%, 81.50\%, 78.85\%, respectively. In contrast, the recognition accuracy of negative and neutral emotions is slightly lower. In addition, the performance difference between positive and neutral emotion of DANN model is 8.69\%, the performance difference between positive and neutral emotion of DAN model is 8.77\%, and the performance difference between positive and negative emotion of DCORAL model is 10.4\%. However, the maximum performance difference of PL-DCP between emotion categories is only 4.09\%. Obviously, PL-DCP showed more robust and balanced performance, while other deep transfer learning methods showed lower stability with fluctuating performance on different emotion categories. 
    On the SEED-IV dataset, as shown in Fig.\ref{fig:confusion_matrix_III_IV}.(e)$\sim$(h), On the SEED-IV dataset, the recognition accuracy of all models for neutral emotion is slightly higher, reaching 76.53\%, 65.96\%, 64.12\% and 61.35\%, respectively, while the recognition accuracy for happy emotion is slightly lower. In addition, on this dataset, each model is easy to confuse sadness and fear. 
    On the SEED-V dataset, as shown in Fig.\ref{fig:confusion_matrix_V}.(a)$\sim$(d), all the models have slightly lower recognition accuracy performance for happy emotions, while they have higher recognition accuracy for neutral, sad, and fear emotions. All models easily confuse happy emotions with fear and neutral emotions. Compared with other models, the PL-DCP model has the best performance in recognizing neutral emotions. Overall, our proposed PL-DCP model has superior emotion recognition performance and stability.

\begin{figure*}[t]
\centering
\includegraphics[width=0.9\textwidth]{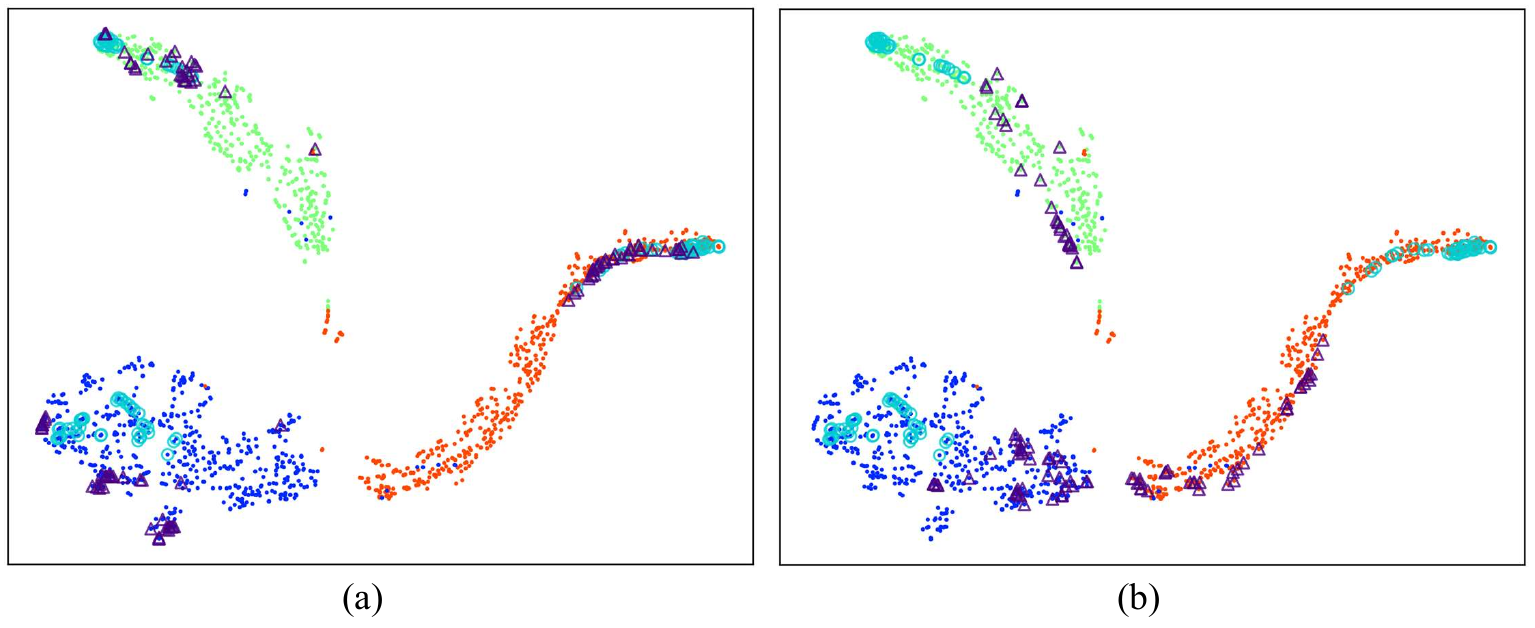}
\caption{A visualization of closer pair (a) and distant pair (b) of domain features in the class prototype space.Here, blue hollow circles ($\circ$) represent domain features from the same subject, and purple hollow triangles ($\bigtriangleup$) represent domain features from another subject. The three clusters with different colors in the figure represent the class feature distribution of the three emotion categories.}
\label{fig:close_fea}
\end{figure*}


\begin{table}[t]
\centering

\caption{Results of ablation experiments with the PL-DCP model, expressed as (Mean-Accuracy$\% \pm$Standard-Deviation\%)}
\color{black}
\label{tab:myseedablation}
\resizebox{\columnwidth}{!}{
\begin{tabular*}{\hsize}{@{\extracolsep{\fill}}lccccc@{}}
\toprule
\multicolumn{3}{c}{\textit{Ablation study about training strategy}} & $P_{acc}(\%)$ \\
\midrule
\multicolumn{3}{l}{w/o domain prototype}                        &74.67±07.62\\
\multicolumn{3}{l}{w/o domain disc. loss}                       &75.24±10.52 \\
\multicolumn{3}{l}{w/o class disc. loss}                        &79.18±06.62 \\
\multicolumn{3}{l}{w/o domain disc. loss and class disc. loss}  &74.49±05.49 \\
\multicolumn{3}{l}{w/o pairwise learning}                       &77.61±09.20\\
\multicolumn{3}{l}{w/o the bilinear transformation matrix S}    &80.64±04.60 \\
\multicolumn{3}{l}{w/o soft regularization R}                   &81.73±03.31 \\
\midrule
\multicolumn{3}{l}{\textbf{\textbf{PL-DCP }}}                      
&\textbf{82.88±05.32} \\
\bottomrule
\end{tabular*}
}
\end{table}

\section{Discussion and Conclusion} 
\label{sec:Discussion_and_Conclusion}
\subsection{Ablation Study}
    \indent To comprehensively evaluate the performance of the model and evaluate the impact of each module in the proposed PL-DCP model, we performed ablation experiments. Ablation results under cross-subject single-session leave-one-subject-out cross-validation based on the SEED dataset are shown in Tab.\ref{tab:myseedablation}. After removing \textit{domain prototype}, it is observed that the performance of the model decreases by 8.21\% when only the class prototype is used, indicating that the domain and class dual prototype representation method has performance advantages compared with the single class prototype representation method. After removing the \textit{domain discriminator loss}, we observe a 7.53\% decrease in model performance. After removing the \textit{class discriminator loss}, the model performance decreases from 82.88\% to 79.18\%, with a decrease of 3.70\%. After removing \textit{domain discriminator loss} and \textit{class discriminator loss}, the model performance decreases significantly by 8.39\%, which indicates that the interaction of domain discriminator and class discriminator helps to extract relevant features and significantly improves the recognition performance for the target domain. After removing the \textit{pairwise learning} strategy and replacing it with the pointwise learning strategy, the model performance decreases by 5.27\%, indicating that the paired learning strategy enhances the recognition performance of the model by capturing the relationship between sample pairs. Removing \textit{bilinear transformation matrix $S$}, the model performance decreases by 2.24\%, indicating that the bilinear transformation matrix $S$ contributes to the model performance in the formula Eq. \ref{Eq:8}. After removing the \textit{soft regularization term $R$}, the model performance decreases by 1.15\%. 
    \\ \indent Overall, All these results demonstrate the effectiveness of the individual components in the PL-DCP model and their combined impact on the overall performance.

\subsection{Visualization of Domain and Class Features}
\label{visualization}
    \indent To intuitively understand the extracted domain and class features, we use T-SNE \cite{2008Tsne} to visualize these features for the respective samples, enabling us a clear observation of how the features and prototypes evolve over training. The visualizations of domain and class features at different training stages are shown in Fig.\ref{fig:fea_tsne}.(a)$\sim$(c) and (d)$\sim$(f), respectively. These figures capture the features at the beginning of training (first column), after 50 training epochs (second column), and at the end of training (third column). 
    \\ \indent In the visualization of domain features from (a)$\sim$(c), different colors represent the domain features for each domain, while diamonds ($\diamond$) in corresponding colors denote the domain prototypes for each domain. The target data are represented as semi-transparent black crosses (×) to avoid excessive overlap with other domain features. Comparing the feature distribution from Fig.\ref{fig:fea_tsne}.(a)$\sim$(c), it is evident that the domain features of the same subject are clustered more and more closely, forming separate groups with the domain prototypes ($\diamond$) located at the center of each cluster. This differentiation in domain feature distributions across subjects supports our hypothesis that domain features, derived through feature disentanglement of shallow features, can effectively distinguish between subjects. A similar trend is observed in the class feature visualization as shown in Fig.\ref{fig:fea_tsne}.(d)$\sim$(f). Here, different colors represent different classes, with the semi-transparent black crosses (×) again indicating the target data. As the training progresses, clearer class boundaries between different categories become more and more obvious, indicating that the model learns a strong feature alignment ability based on class prototypes during training.
    \\ \indent To further illustrate the relationships between domain prototypes and class prototypes, we analyze both close pairs and distant pairs of domain samples. As shown in Fig.\ref{fig:close_fea}.(a)$\sim$(b), we visualizing their respective representations in the class prototype space.(a) shows that the domain features of the subject pairs are relatively close, with a high degree of overlap, and the class feature distributions are therefore relatively close. (b) shows that the domain features of the subject pairs are relatively far away, with a low degree of overlap, and the class feature distributions are therefore relatively far away. The results reveal that Samples that are close to each other in the domain prototype space also tend to remain close in the class prototype space, indicating consistency and coherence in the mapping across the two prototype spaces. This observation reinforces the effectiveness of the proposed framework in preserving the intrinsic relationships between samples during dual prototype learning.

\subsection{Effect of Noisy Labels}
\label{noisylabels}
    \indent We further evaluate the performance of the model under label noise to evaluate the robustness and noise immunity introduced by pairwise learning. During the experiment, the labels of $\eta$\% (5\%, 10\%, 20\% and 30\%) of the source domain data are randomly replaced with random error labels, which simulates the situation that the data labels contain noise in the real scene. We tested the PL-DCP model using cross-subject single-session leave-one-out cross validation on the SEED database, and compared it on pointwise learning and pairwise learning strategies.
    \\ \indent The results are shown in Tab.\ref{tab:noisy labels}. When using the pointwise learning strategy, The accuracy of the model in different proportions of label noise was 77.61\%, 76.76\%, 73.49\%, 69.17\%, 66.53\%. The performance of the model decreases rapidly with the increase of $\eta$, and the performance of the model decreases by 11.08\% when the label noise rate reaches 30\%. When using pairwise learning strategy, the accuracy rates are 81.46\%, 80.32\%, 79.79\% and 79.01\%, respectively. As the label noise rate gradually increases from 5\% to 30\%, the overall performance of the model only decreases by 3.87\%. The results show that the performance of the PL-DCP model only slightly decreases in the label noise environment, showing superior performance, and the introduction of pairwise learning strategy shows strong robustness under label noise.
    
\begin{table}[]
\begin{center}
\caption{Results of the PL-DCP model adding different proportions($\eta\%$) of label noise to the source domain, expressed as (Mean-Accuracy$\% \pm$Standard-Deviation\%).}
\label{tab:noisy labels}
\color{black}
\begin{tabular}{ccc}
\toprule
Noisy Ratio   & Pointwise Learning      & Pairwise Learning  \\ 
$\eta\ (\%)$      &   $P_{acc}(\%)$         &     $P_{acc}(\%)$  \\
\midrule
0\%           &  77.61$\pm$09.20     &  82.88$\pm$05.32 \\
5\%           &  76.76$\pm$08.64     &  81.46$\pm$05.54 \\
10\%          &  73.49$\pm$07.96     &  80.32$\pm$06.39 \\
20\%          &  69.17$\pm$08.31     &  79.79$\pm$06.09 \\
30\%          &  66.53$\pm$06.63     &  79.01$\pm$07.46 \\
\bottomrule
\end{tabular}
\end{center}
\end{table}
    
\subsection{Conclusion}
    \indent This study proposes a novel pairwise learning framework with domain and class prototypes (PL-DCP) for EEG-based emotion recognition in unseen target conditions. Unlike existing transfer learning methods that require both source and target data for feature alignment, PL-DCP relies solely on source data for model training. Experimental results show that the proposed method achieves promising results even without using target domain data for training, with performance approaching or even surpassing some deep transfer learning models that heavily rely on target domain data. This suggests that combining feature disentanglement with domain and class prototypes helps generalize more reliable and stable characteristics of individual subjects. Additionally, the introduction of pairwise learning enhances the anti-interference ability of the model to label noise. These findings underscore the potential of this method for practical applications inAffective Brain-Computer Interfaces(aBCIs). In the future work, we will further explore aBCIs models with better performance and more adaptability.

\section*{Acknowledgements}
    \indent This work was supported in part by the National Natural Science Foundation of China under Grant 62176089, 62276169 and 62201356, in part by the Natural Science Foundation of Hunan Province under Grant 2023JJ20024, in part by the Key Research and Development Project of Hunan Province under Grant 2025QK3008, in part by the Key Project of Xiangjiang Laboratory under Granted 23XJ02006, in part by the STI 2030-Major Projects 2021ZD0200500, in part by the Medical-Engineering Interdisciplinary Research Foundation of Shenzhen University under Grant 2024YG008, in part by the Shenzhen University-Lingnan University Joint Research Programme, and in part by Shenzhen-Hong Kong Institute of Brain Science-Shenzhen Fundamental Research Institutions (2023SHIBS0003). 
\appendix




\bibliographystyle{elsarticle-num} 
\bibliography{references}





\end{document}